\documentclass[preprint,12pt]{elsarticle}

\usepackage{amssymb}
\usepackage{amsmath}
\usepackage{booktabs}
\usepackage{multirow}
\usepackage{amssymb}
\usepackage{array}
\usepackage{makecell}

\usepackage[hidelinks]{hyperref}
\journal{Knowledge-Based Systems}

\begin{document}

\begin{frontmatter}



\title{Skeleton-Guided Sparse Anchors for Rotated Instance Segmentation in Cell Microscopy}


\author[HZCU]{Jun Wang\fnref{associated}}
\ead{wangjun@hzcu.edu.cn}

\author[AIER]{Chengfeng Zhou\fnref{associated}}
\ead{joe1chief1993@gmail.com}

\author[HZCU]{Zhaoyan Ming}
\ead{mingzhaoyan@gmail.com}

\author[HZCU]{Lina Wei}
\ead{weiln@hzcu.edu.cn}

\author[FDU]{Songchang Chen}
\ead{26775172@qq.com}

\author[NTU]{Xudong Jiang}
\ead{exdjiang@ntu.edu.sg}

\author[FDU]{Chenming Xu}
\ead{xuchenm@163.com}

\author[SJTU]{Dahong Qian\corref{cor1}}
\ead{dahong.qian@sjtu.edu.cn}

\fntext[associated]{J. Wang and C. Zhou contributed equally to this work as co-first authors.}
\cortext[cor1]{Corresponding author.}

\affiliation[HZCU]{organization={School of Computer and Computing Science, Hangzhou City University},
            city={Hangzhou},
            postcode={310015}, 
            country={China}}

\affiliation[AIER]{organization={Aier Institute of Digital Ophthalmology and Visual Science, Changsha Aier Eye Hospital},
            city={Changsha},
            postcode={410000}, 
            country={China}}

\affiliation[SJTU]{organization={School of Biomedical Engineering, Shanghai Jiao Tong University},
            city={Shanghai},
            postcode={200030}, 
            country={China}}

\affiliation[NTU]{organization={Department of Electrical and Electronic Engineering, Nanyang Technological University},
            city={Singapore},
            postcode={639798}, 
            country={Singapore}}

\affiliation[FDU]{organization={Obstetrics and Gynecology Hospital, Institute of Reproduction and Development, Fudan University},
            city={Shanghai},
            postcode={200011}, 
            country={China}}

\begin{abstract}
One of the fundamental challenges in cell microscopy (MS) image analysis is instance segmentation (IS), particularly when segmenting cluster regions where multiple objects of varying sizes and shapes may be connected or even overlapped in arbitrary orientations. Existing IS methods usually fail in handling such scenarios, as they rely on coarse instance representations such as keypoints and horizontal bounding boxes (h-bboxes). In this paper, we propose a novel one-stage framework named A2B-IS to address this challenge and enhance the accuracy of IS in MS images. Our approach represents each instance with a pixel-level mask map and a rotated bounding box (r-bbox). Unlike two-stage methods that use box proposals for segmentations, our method decouples mask and box predictions, enabling simultaneous processing to streamline the model pipeline. Additionally, we introduce a Gaussian skeleton map to aid the IS task in two key ways: (1) It guides anchor placement, reducing computational costs while improving the model’s capacity to learn RoI-aware features by filtering out noise from background regions. (2) It ensures accurate isolation of densely packed instances by rectifying erroneous box predictions near instance boundaries. To further enhance the performance, we integrate two modules into the framework: (1) An Atrous Attention Block (A2B) designed to extract high-resolution feature maps with fine-grained multiscale information, and (2) A Semi-Supervised Learning (SSL) strategy that leverages both labeled and unlabeled images for model training. Our method has been thoroughly validated on two large-scale MS datasets, demonstrating its superiority over most state-of-the-art approaches.

\end{abstract}




\end{frontmatter}


\section{Introduction}
\label{sec:introduction}
\begin{figure*}[!t]
    \centering
    \includegraphics[width=1.0\textwidth]{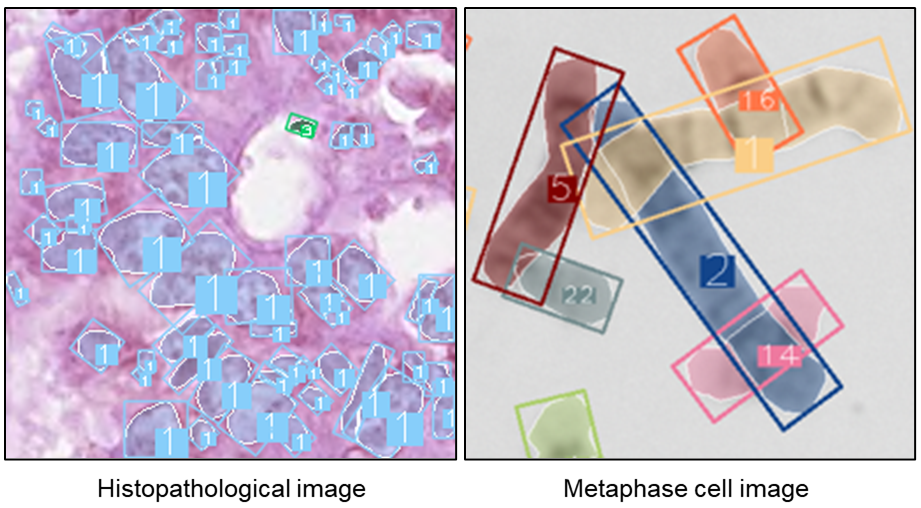}

    \caption{Instance segmentation in microscopy images is a highly challenging task due to the special characteristics of objects, e.g., irregular shapes, small but diverse sizes, and dense distribution (even cross-overlapped) in arbitrary orientations. For clarity, we only show a local zoom region of each MS image, and the numbers in the boxes indicate the class labels.}
\label{fig:1}
\end{figure*}

Microscopy (MS) image is widely used in clinical practice as a golden standard tool for the diagnosis of various human diseases, e.g., cancers and chromosome disorders. Instance segmentation (IS) plays a crucial role in MS image analysis. For example, nuclei segmentation in histopathological images is a prerequisite for analyzing the tumor microenvironment~\citep{citiation1}. Chromosome segmentation in metaphase cell images is an essential step for karyotyping~\citep{citiation2}. As demonstrated in Fig.~\ref{fig:1}, one fundamental challenge of IS in MS images is to separate each instance in crowded regions, where multiple objects of varying sizes and shapes may be inter-connected or even cross-overlapped in arbitrary orientations.

\begin{figure*}[!t]
    \centering
    \includegraphics[width=1.0\textwidth]{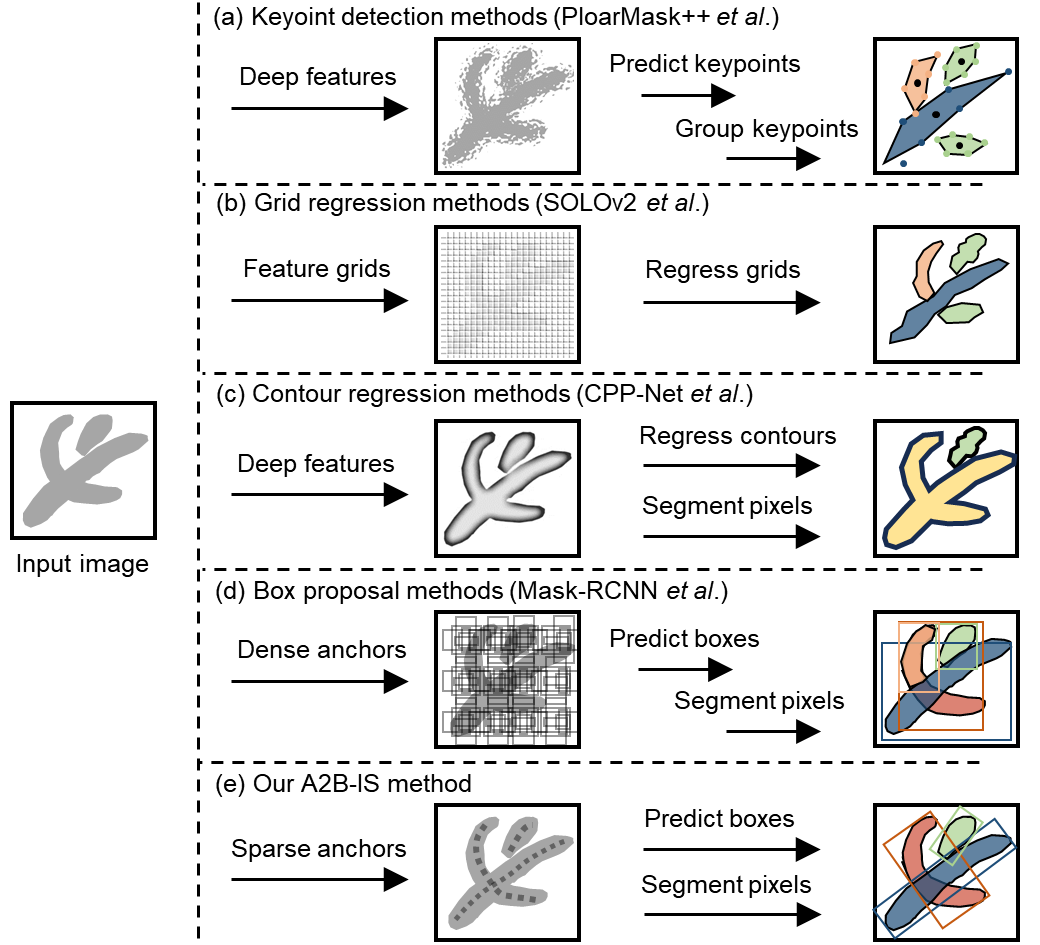}

    \caption{Illustration of various instance separation strategies. One-stage methods represent instances typically using (a) key points or (b) grid cells, which are too coarse for instance identification. Two-stage methods like (c) Mask-RCNN contain many redundant anchors and may fail to recognize objects in cluster regions with horizontal boxes. (d) Our A2B-IS method separates instances directly using pixel-wise masks and rotated boxes in a single stage. The rotated boxes are predicted from skeleton-guided sparse anchors.}
\label{fig:2}
\end{figure*}

Instance separation strategies employed by existing IS methods can be roughly summarized into three categories: keypoint detection, grid regression, and box proposal as illustrated in Fig.~\ref{fig:2}(a)-(c). (1) The keypoint methods~\citep{citiation3,citiation4} first detect instance-agnostic keypoints, then group them into corresponding instances in a post-processing stage. For example, the method PolarMask++~\citep{citiation3} represents instances using centroids and ray lengths in the polar coordinate (sampled contour points). (2) The grid regression methods, e.g., the SOLO~\citep{citiation5} and SOLOv2~\citep{citiation6}, directly predict instance masks based on image grids without any post-processing. (3) The box proposal methods normally follow a two-stage framework such as the Mask-RCNN~\citep{citiation7} that segments instances based on a proposal of horizontal bounding boxes (h-bboxes). 

Although the above methods have achieved remarkable results in their respective tasks, they may fail to handle the unique characteristics of objects in MS images due to their inappropriate instance representations, i.e., the h-bboxes, grids, and keypoints. As illustrated in Fig.~\ref{fig:2}, these representations cannot accurately identify instances within cluster regions. The one-stage methods often produce poor masks and are unable to handle the overlap issue since they separate objects using coarse information such as centroids and grids. In contrast, the two-stage methods can achieve more fine-gained pixel-level masks and have the potential to isolate overlapped instances with bounding boxes. Nevertheless, they also suffer from limitations such as complicated model pipeline and dense anchor strategy. In particular, the dense anchor strategy leads to numerous redundant anchors and involves extra hyperparameters, e.g., anchor base sizes and aspect ratios, that must be meticulously tuned for specific tasks to guarantee promising performance. 

Moreover, most existing methods follow the high-to-low (encoder) and low-to-high (decoder) feature extraction paradigm, such as the UNet~\citep{citiation8} and FPN~\citep{citiation9}. This can result in spatial information loss and adversely affect the segmentation performance, which becomes more severe for densely packed small objects in MS images, especially those with slender shapes. The HRNet~\citep{citiation10} alleviates this issue by maintaining high-resolution feature maps throughout the whole feature extraction process. Nevertheless, this design involves repeated information exchange across different resolutions to perceive multiscale semantic representations, leading to a complicated model pipeline, and resolution loss persisting.

Furthermore, training a robust deep model for the above-mentioned methods typically relies on a large quantity of densely annotated images. However, labeling MS images can be expensive and time-consuming due to the expertise required to accurately annotate these images. For example, we collected more than 4,000 metaphase cell images for the chromosome segmentation task in two weeks. Despite the dedicated effort of five experienced cytogeneticists, it took approximately eight months to annotate only 615 (15.4\%) images. Fortunately, Semi-Supervised Learning (SSL) can adopt a “teacher-student” architecture to mine supervision signals from the unlabeled images via pseudo-labeling or consistent regularization. Although the effectiveness of SSL has been widely validated in image classification~\citep{citiation11,citiation12} and semantic segmentation problems~\citep{citiation13}, conducting the SSL for the aforementioned IS methods in MS images is nontrivial due to their weak instance representations and complex model pipeline.

To address the distinctive challenges posed by MS images, this paper proposes a novel method named A2B-IS for instance segmentation in MS images. Compared to existing methods, our A2B-IS has three main advantages. Firstly, as illustrated in Fig.~\ref{fig:2}(d), our method represents instances with pixel-level masks and rotated bounding boxes (r-bboxes), which can more accurately separate objects in MS images. Unlike two-stage methods that perform segmentations relying on box-proposals, our mask and box predictions are decoupled and performed simultaneously in a single-stage, leading to a simpler and more efficient pipeline. Additionally, a Gaussian skeleton map is introduced to aid the IS task in two key ways: (1) It guides anchor placement on the foreground regions (see Fig. 2d), which \textit{not only} saves computational costs through reducing redundant anchors \textit{but also} enhances the model’s capacity to learn RoI-aware features by filtering out noise from background regions. (2) It ensures accurate isolation of densely packed instances by rectifying erroneous box predictions near the instance boundaries (see Section~\ref{sec:box-mask-proposal} for more details).

Secondly, considering that a high-resolution feature map is vital for the MS image analysis, we designed a novel Convolutional Neural Network (CNN)-based module named Atrous Attention Block (A2B) to construct the backbone network. This module can help even a small model to extract high-resolution feature maps containing fine-grained multiscale information, further improving the segmentation performance.  

Finally, benefiting from the simplified model pipeline and enhanced instance representations, we design a Semi-Supervised Learning strategy for IS (SSL-IS) that transforms the IS task into a manageable semantic segmentation problem. This strategy generates pixel-wise pseudo labels for unlabeled images instead of RoI-level labels, making it easy to extend existing SSL methods for semantic segmentation tasks into the IS domain. Based on this strategy, abundant unlabeled images can be leveraged to train the model and further boost its performance.

To verify the proposed method, extensive experiments are conducted on two large-scale representative datasets: a public dataset named \textit{PanNuke}~\citep{citiation14} for nuclei segmentation and a private dataset named \textit{ChromSeg-SSL} for chromosome segmentation. The experimental results demonstrate that the proposed A2B-IS can achieve much superior performance to other methods on MS images. Overall, our main contributions can be summarized in four aspects as follows:

\begin{enumerate}
    \item \textbf{A2B-IS}: We introduce a pioneering one-stage instance segmentation method called A2B-IS, specifically designed to tackle the unique challenges presented by microscopy images. This novel approach effectively handles objects with dense distribution in arbitrary orientations via a skeleton-guided instance representation strategy. 

    \item \textbf{Atrous Attention Block (A2B)}: We present a CNN-based module, the A2B, which excels at learning high-resolution feature maps containing crucial multiscale information. This module is essential for the fine-grained instance segmentation of microscopy images.

    \item \textbf{SSL-IS Strategy}: We develop a powerful Semi-Supervised Learning Instance Segmentation (SSL-IS) strategy that allows for the seamless adaptation of cutting-edge SSL methods initially designed for classification or semantic segmentation tasks into the instance segmentation domain. This strategy integration could leverage unlabeled images to further enhance the segmentation performance.

    \item \textbf{ChromSeg-SSL Dataset}: To facilitate research in the field of microscopy image analysis, we release a comprehensive large-scale dataset named ChromSeg-SSL. This dataset comprises 4,185 metaphase cell images with a resolution of 1600$\times$1600 pixels. Notably, 615 images have been annotated by five experienced cytologists. We envision that this dataset will significantly contribute to the advancement of microscopy image analysis research.
    
\end{enumerate}

\section{Related Work}
\label{sec:related_work}

\subsection{Instance Segmentation for Natural Images}
The Mask-RCNN~\citep{citiation7} pioneers deep-learning-based instance segmentation methods, which predict masks relying on box proposals. Most following methods, such as the Cascade Mask-RCNN~\citep{citiation15}, HTC~\citep{citiation16}, SCNet~\citep{citiation17}, and MS-RCNN~\citep{citiation18}, extend the Mask-RCNN with different strategies to learn more discriminative features. These methods are still prominent in various instance segmentation tasks, as they can guarantee SOTA performance. However, the pipeline of the Mask-RCNN structure is complicated and suffers from high computational costs.

Recently, single-stage methods~\citep{citiation3,citiation6} also developed rapidly and achieved appealing results at a faster speed. They aim to predict masks directly, avoiding dependence on box proposals. However, existing single-stage methods represent instances either based on grids, centroids, contours, or h-bboxes, which may limit their performance when directly applied to MS images, especially for chromosome instance segmentation. Objects such as chromosomes in MS images have arbitrary orientations, slender and bent shapes, and may even cross-overlapped with each other. In this case, grids, centroids, contours, or h-bboxes are unable to effectively separate these objects. In contrast, our proposed A2B-IS represents objects using pixel-wise masks and skeleton-guided r-bboxes, which is more suitable for instance segmentation in MS images.

\subsection{Instance Segmentation for Microscopy Images}
Some prior studies also have tried to address the instance segmentation problem in MS images, with mostly focused on the nuclei segmentation task~\citep{citiation1,citiation4,citiation19,citiation20,citiation21,citiation22}. For instance, Liu et al.~\citep{citiation23} proposed a CNN-based network named PFFNet for biomedical image segmentation. Graham et al.~\citep{citiation20} developed the Hover-Net for nuclei segmentation in histology images. Wang et al.~\citep{citiation24} modified the Mask-RCNN for chromosome instance segmentation in metaphase cell images. While these methods can produce impressive results on their respective tasks, they still fall into the reliance on predictions of centroids, contours or h-bboxes. In addition, achieving top performance in these methods requires a substantial number of pixel-level annotated images which are quite costly to obtain.

\subsection{Semi-Supervised Learning}
SSL methods represent highly effective solutions for addressing the scarcity of annotations. These methods can be roughly classified into two categories: consistent regularization and pseudo-labeling. Regularization-based methods~\citep{citiation25} establish a loss function to ensure that predictions made under a set of perturbations are consistent. On the other hand, the pseudo-labeling-based methods~\citep{citiation13} are more straightforward. They initially train a model using labeled data and then regard the pre-trained model as a teacher to generate pseudo-labels for unlabeled data. Finally, all data are combined to train a student model. However, the above training pipeline is troublesome. Generally, the teacher model is implemented by using the student’s Exponential Moving Average (EMA) to perform online pseudo-labeling~\citep{citiation11}. 
Even though many teacher-student based SSL methods such as Mean-Teacher~\citep{citiation11} and Fix-Match~\citep{citiation25} have been developed for image classification or semantic segmentation tasks, it remains challenging to extend these approaches to IS problems, particularly in MS images. Existing studies attempted to design complicated SSL techniques for specific IS tasks in MS images. For instance, Zhou et al.~\citep{citiation26} have proposed a deep semi-supervised knowledge distillation method for overlapping cervical cell instance segmentation. Liu et al.~\citep{citiation27} trained the Mask-RCNN with an unsupervised domain adaptation method for cell instance segmentation in histopathology images. However, they are complex and highly sensitive to hyperparameters~\citep{citiation28}, since they rely on the two-stage-box-proposal structure.

\begin{figure*}[!t]
    \centering
    \includegraphics[width=1.0\textwidth]{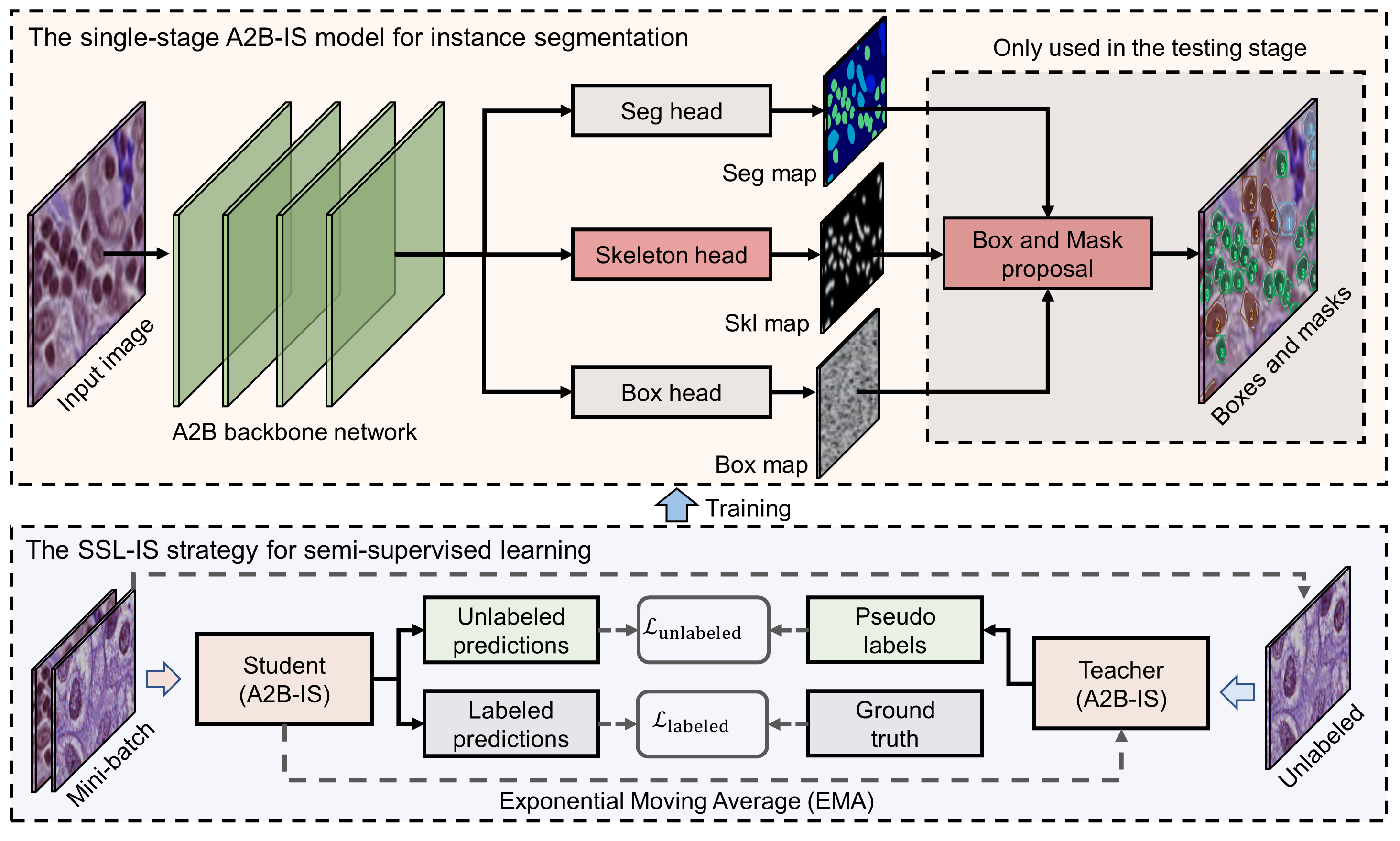}

    \caption{The framework of the proposed A2B-IS method for instance segmentation of microscopy images. High-resolution features are extracted from the input image using the A2B backbone network. Then, the features are fed to three head subnetworks for predicting object regions, object skeletons, and box transformation parameters. Finally, the predictions propose the box and mask for each instance. Noting that box and mask proposal is only needed in the testing stage, which facilitates the SSL-IS that leverages the unlabeled images to train the model.}
\label{fig:3}
\end{figure*}

\section{Method}
\label{sec:method}
Fig.~\ref{fig:3} illustrates the main architecture of the proposed A2B-IS method. High-resolution features are first extracted from the input image using the A2B backbone network. Then, the features are fed to three subnetworks (i.e., the Seg head, the Skeleton head, and the Box head in Fig.~\ref{fig:3}) for predicting object masks, skeletons, and box transformation parameters, respectively. These predictions are finally utilized to obtain the boxes and masks for each instance. The key components of the framework are three-fold: 1) the A2B block for building the backbone network, 2) the SSL-IS strategy for training the model using both labeled and unlabeled data, and 3) the box and mask proposal module in the testing stage for generating the final results. Details are presented in the following subsections.

\begin{figure*}[!t]
    \centering
    \includegraphics[width=1.0\textwidth]{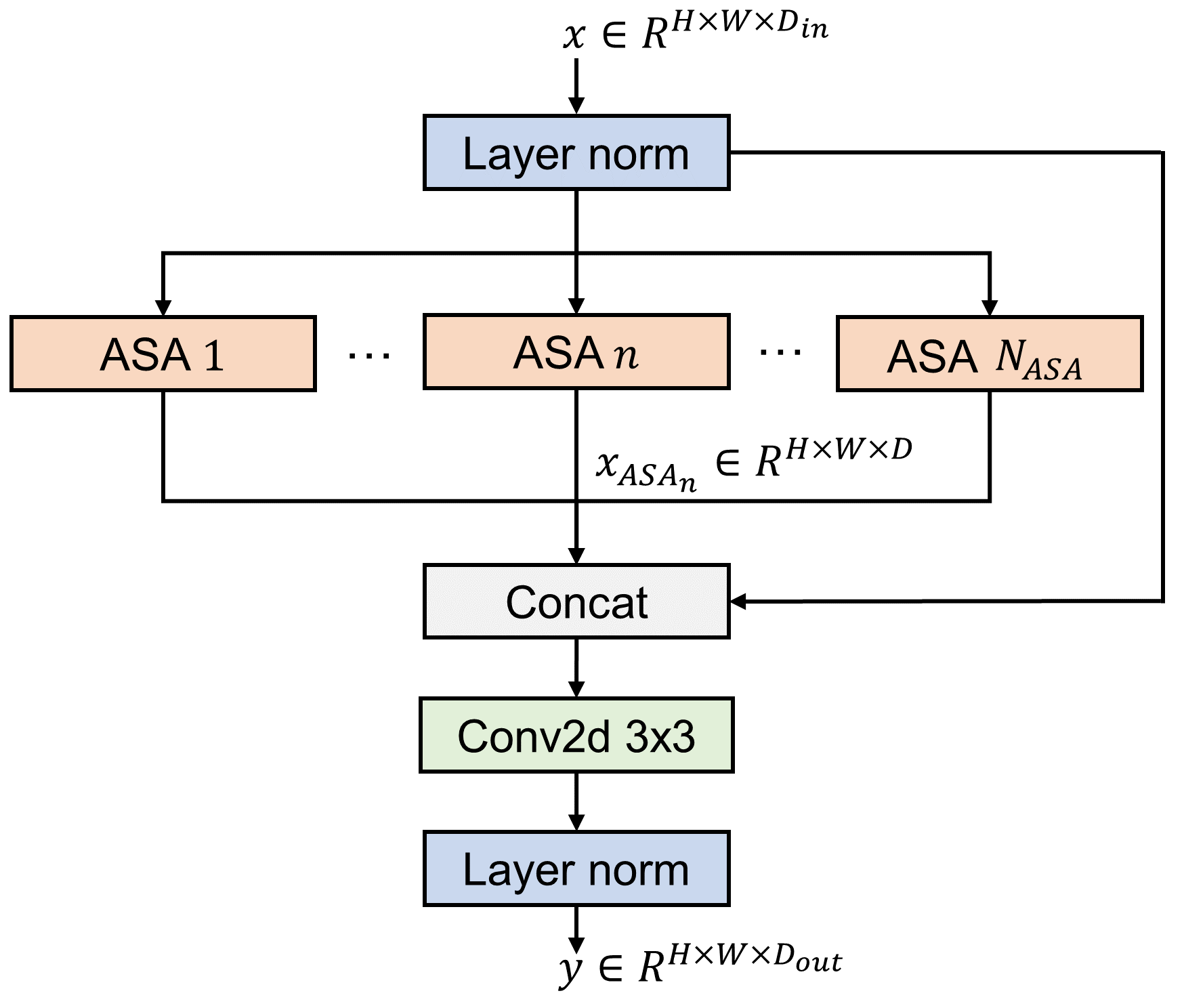}
    \caption{The proposed A2B block. It contains multiple parallel Atrous Self-Attention (ASA) modules to extract multiscale highly discriminative features.}
\label{fig:4}
\end{figure*}

\subsection{A2B Block for Building Backbone Network}
The backbone network is built via cascading several A2B blocks to extract high-resolution feature maps from the input MS image. Fig.~\ref{fig:4} illustrates the structure of an A2B block. Let the input and output feature maps of the block be $\boldsymbol{x} \in \mathbb{R}^{H \times W \times D_{\text{in}}}$ and $\boldsymbol{y} \in \mathbb{R}^{H \times W \times D_{\text{out}}}$, respectively, the mapping relationship can be expressed as follows:
\begin{equation}
    \boldsymbol{y} = \text{LN}\left( \text{CV}\left([\boldsymbol{x}_{\text{norm}}, \boldsymbol{x}_{\text{ASA}_{n}} \mid n = 1, \dots, N_{\text{ASA}}]\right) \right),
    \label{eq:1}
\end{equation}
where $\text{CV}$ and $\text{LN}$ indicate the convolutional layer and the layer normalization~\citep{citiation29}, respectively. $\boldsymbol{x}_{\text{norm}}$ is the normalized feature map of the input $\boldsymbol{x}$. The term $[\cdot]$  denotes the concatenation of the $\boldsymbol{x}_{\text{norm}}$ and the outputs $\boldsymbol{x}_{\text{ASA}}$ of $N_{\text{ASA}}$ parallel Atrous Self-Attention ($\text{ASA}$) modules. The $n_{th}$ module is defined as:
\begin{equation}
    \boldsymbol{x}_{\text{ASA}n} = \text{LN}\left(v_n \ast \text{Softmax}\left(\boldsymbol{k}_n \ast \boldsymbol{q}_n\right)\right),
    \label{eq:2}
\end{equation}
where $\ast$ denotes the pixel-wise multiplication, while $\boldsymbol{k}=k(\boldsymbol{x}_{\text{norm}})$, $\boldsymbol{q}=q(\boldsymbol{x}_{\text{norm}})$, and $\boldsymbol{v}=v(\boldsymbol{x}_{\text{norm}})$ are the key, query, and value of the $\boldsymbol{x}_{\text{norm}}$, respectively. Unlike the Transformer~\citep{citiation30} which uses the fully connected layers to implement the $k$, $q$, and $v$, we adopt the atrous convolution to perform these operations, since it is more efficient for high-resolution inputs. Formally, these operations can be formulated as:
\begin{equation}
    \begin{aligned}
    \boldsymbol{k}_n &= \text{AtrousCV}\left(f(\boldsymbol{x}_{\text{norm}}), \text{rate} = n \right), \\
    \boldsymbol{q}_n &= \text{AtrousCV}\left(f(\boldsymbol{x}_{\text{norm}}), \text{rate} = n \right), \\
    \boldsymbol{v}_n &= \text{AtrousCV}\left(f(\boldsymbol{x}_{\text{norm}}), \text{rate} = n \right), \\
    \end{aligned}
    \label{eq:3}
\end{equation}
where $\text{rate} = n$ is the dilation rate of the atrous convolution. $f$ is an optional transformation function, e.g., a convolutional layer for down-sampling in the channel direction to reduce computational costs. The term $\text{Softmax}\left(\boldsymbol{k}_n \ast \boldsymbol{q}_n\right)$ in Eq.~\ref{eq:2} denotes the attention map. Let $\boldsymbol{M} = \boldsymbol{k}_n \ast \boldsymbol{q}_n \in \mathbb{R}^{H \times W \times D}$, the attention map be calculated as follows:
\begin{equation}
    x^{ijc}_{\text{attn}} = \frac{e^{M_{ijc}}}{\sum_{(i', j') \in \Omega} e^{M_{i'j'c}}}, 
    \label{eq:4}
\end{equation}
where $(i,j,c)$ is the pixel index, and $\Omega$ denotes the spatial space of the map. According to Eq.~\ref{eq:4}, the softmax is performed on the spatial space rather than the channels. Since the A2B block contains multiple parallel ASA modules with different dilation rates, it can extract highly discriminative features with multiscale information.

\subsection{The SSL-IS Strategy for Model Training}

\begin{figure*}[!t]
    \centering
    \includegraphics[width=1.0\textwidth]{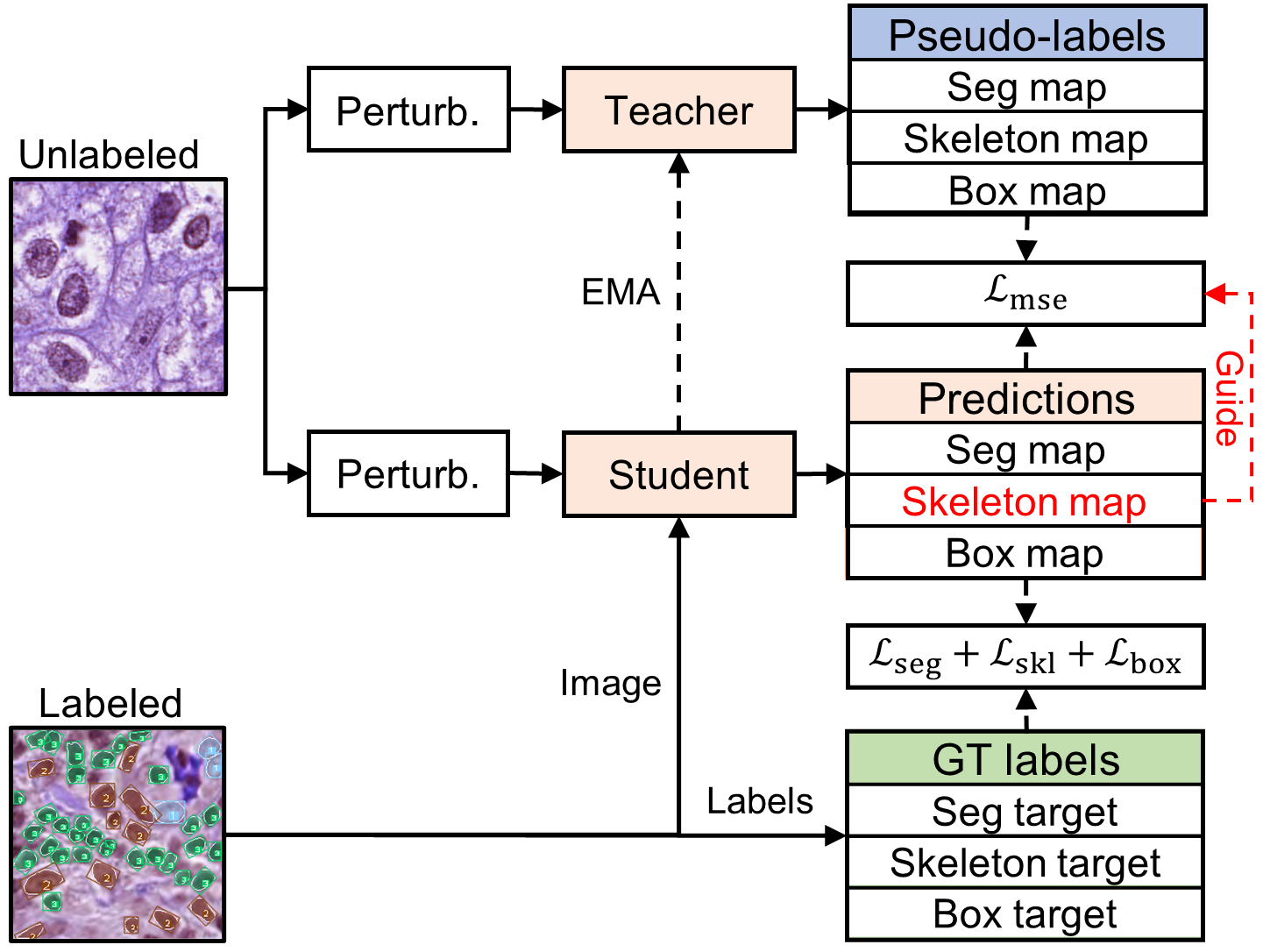}
    \caption{The proposed SSL-IS strategy can use unlabeled data for model training. The teacher model, which served as an Exponential Moving Average (EMA) of the student, generates pseudo-labels for the unlabeled data. The skeleton map determines the foreground regions for distilling the pseudo-supervision.}
\label{fig:5}
\end{figure*}

Our proposed A2B-IS model is like a semantic segmentation method so that any SSL methods~\citep{citiation11,citiation25} for semantic segmentation can be extended to train our model. In this section, we propose an SSL-IS strategy that follows the Mean-Teacher structure~\citep{citiation11}. Fig.~\ref{fig:5} shows the main idea of the strategy. For the labeled data, the student is trained via supervised learning between its three predictions (i.e., the maps generated by the three heads) and the ground-truth (GT) targets. For the unlabeled data, the teacher model, as an EMA of the student model, is utilized to generate online pseudo-labels for the student’s predictions. Then, a consistent regularization loss is adopted to supervise learning between the pseudo-labels and the student’s predictions. To enhance the generalization ability of the model, the image is first randomly perturbated before feeding it to the teacher and the student. In this study, perturbations are performed using simple image processing techniques, including random brightness and random contrast.

Let us denote the student’s predictions as $M_{\text{skl}}^s \in \mathbb{R}^{H \times W \times 1}
$, $M_{\text{seg}}^s \in \mathbb{R}^{H \times W \times C}
$, and $M_{\text{box}}^s \in \mathbb{R}^{H \times W \times 5}
$. The skeleton map $M_{\text{skl}}^s$ predicts sigmoid values that estimate the Gaussian distributions of each chromosome’s skeleton points. The segmentation map $M_{\text{seg}}^s$ predicts softmax probabilities that determine the regions of $C=N_{\text{cls}}+2$ categories, where $N_{\text{cls}}$ denotes the number of object categories, and the number $2$ indicates two additional channels that are used to predict the background and the overlapped regions, respectively. The box map $M_{\text{box}}^s$ predicts five offset items used to move, scale, and rotate anchors for accurate localization of objects. 

Notably, we set only a single anchor of size $3\times3$ at each pixel location in the foreground regions. These regions are determined via thresholding of the skeleton map: $M_{\text{skl}}^s\geq \delta$, where $\delta$ is empirically set to $0.02$ in this study. This skeleton-guided single-anchor strategy contains almost no anchor-related hyperparameters and saves computational costs by reducing redundant anchors. More importantly, given that a large anchor may encompass multiple partial or intact objects in crowded regions, we opt for a small anchor (i.e., $3\times3$) to suppress the dense distribution issue.

Let the teacher’s pseudo skeleton map, the segmentation map, and the box map be $M_{\text{skl}}^t \in \mathbb{R}^{H \times W \times 1}
$, $M_{\text{seg}}^t \in \mathbb{R}^{H \times W \times C}
$, and $M_{\text{box}}^t \in \mathbb{R}^{H \times W \times 5}
$, respectively, the following Mean Squire Error (MSE) loss is adopted to train the model:
\begin{equation}
    \mathcal{L}_{\text{unlabeled}} = \text{MSE}(M_{\text{cat}}^t, M_{\text{cat}}^s),
    \label{eq:5}
\end{equation}
where $M_{\text{cat}}^t = \left[ M_{\text{skl}}^t, M_{\text{seg}}^t, M_{\text{box}}^t \right]
$ is the concatenation of the teacher’s pseudo maps along the channels, and $M_{\text{cat}}^s$
is the concatenation of the student’s predictions, i.e., $\left[ M_{\text{skl}}^s, M_{\text{seg}}^s, M_{\text{box}}^s \right]$. Notably, to reduce the interference from background regions, we activate the loss calculations only on the foreground regions. 

Concerning the labeled data, we denote the GT skeleton map, the segmentation map, and the box map as $M_{\text{skl}}^g \in \mathbb{R}^{H \times W \times 1}
$, $M_{\text{seg}}^g \in \mathbb{R}^{H \times W \times C}
$, and $M_{\text{box}}^g \in \mathbb{R}^{H \times W \times 5}$, respectively. Then, the following multi-task loss function is adopted to train the model:
\begin{equation}
    \mathcal{L}_{\text{labeled}} = \mathcal{L}_{\text{skl}} + \mathcal{L}_{\text{seg}} + \mathcal{L}_{\text{box}},
    \label{eq:6}
\end{equation}
where $\mathcal{L}_{\text{skl}}$ is the Quality Focal Loss~\citep{citiation31} as follows:
\begin{equation}
    \mathcal{L}_{\text{skl}} = -\left| M_{\text{skl}}^g - M_{\text{skl}}^s \right|^\gamma \ast \left[ M_{\text{skl}}^g \log(M_{\text{skl}}^s) + (1 - M_{\text{skl}}^g) \log(1 - M_{\text{skl}}^s) \right], 
    \label{eq:7}
\end{equation}
where $\gamma=2$ is the suppression factor. The $\mathcal{L}_{\text{seg}}$ in Eq.~\ref{eq:6} is defined as follows:
\begin{equation}
    \mathcal{L}_{\text{seg}} = \mathcal{L}_{\text{ce}}(M_{\text{seg}}^g, M_{\text{seg}}^s) + \mathcal{L}_{\text{dice}}(M_{\text{seg}}^g, M_{\text{seg}}^s), 
    \label{eq:8}
\end{equation}
where $\mathcal{L}_{\text{ce}}$ and $\mathcal{L}_{\text{dice}}$ denote the Cross-Entropy and the Dice loss, respectively.

The box loss $\mathcal{L}_{\text{box}}$ in Eq.~\ref{eq:6} is implemented as the Kullback-Leibler Divergence Loss~\citep{citiation32}:
\begin{equation}
    \mathcal{L}_{\text{box}} = \frac{1}{|S_{\text{ach}}|} \sum_{i \in S_{\text{ach}}} \left[ 1.0 - \frac{1.0}{\tau + \ln(D_{i} + 1.0)} \right],
    \label{eq:9}
\end{equation}
where $S_{\text{ach}}$ is the set of anchors that are set on the foreground regions, and $i$ is the anchor index.  $\tau=1$ is a hyperparameter used to modulate the box loss.  $D_{i}$ is the Kullback-Leibler divergence between the Gaussian distribution of the $i_{\text{th}}$ box prediction (decoded from the anchor based on the $M_{\text{box}}^s$) and the Gaussian distribution of its GT box. Formally, $D$ is calculated as follows:
\begin{equation}
    D = \left\| \mu_p - \mu_g \right\|_2^2 + \text{Tr}(\sigma_p + \sigma_g - \sqrt{\sqrt{\sigma_p} \sigma_g \sqrt{\sigma_p}}), 
    \label{eq:10}
\end{equation}
where $\mu=(x,y)^{T}$ and $\sigma = \left( \begin{matrix}
\frac{w}{2} \cos^2\theta + \frac{h}{2} \sin^2\theta & \frac{w - h}{2} \cos\theta \sin\theta \\
\frac{w - h}{2} \cos\theta \sin\theta & \frac{w}{2} \sin^2\theta + \frac{h}{2} \cos^2\theta
\end{matrix} \right)
$ is the expectation and variance of 2D Gaussian distribution corresponding to an arbitrary-oriented bounding box that denotated by $(x,y,w,h,\theta)$. The subscripts $p$ and $g$ in Eq.~\ref{eq:10} denote the predicted bounding box and GT bounding box, respectively.

\begin{figure*}[!t]
    \centering
    \includegraphics[width=1.0\textwidth]{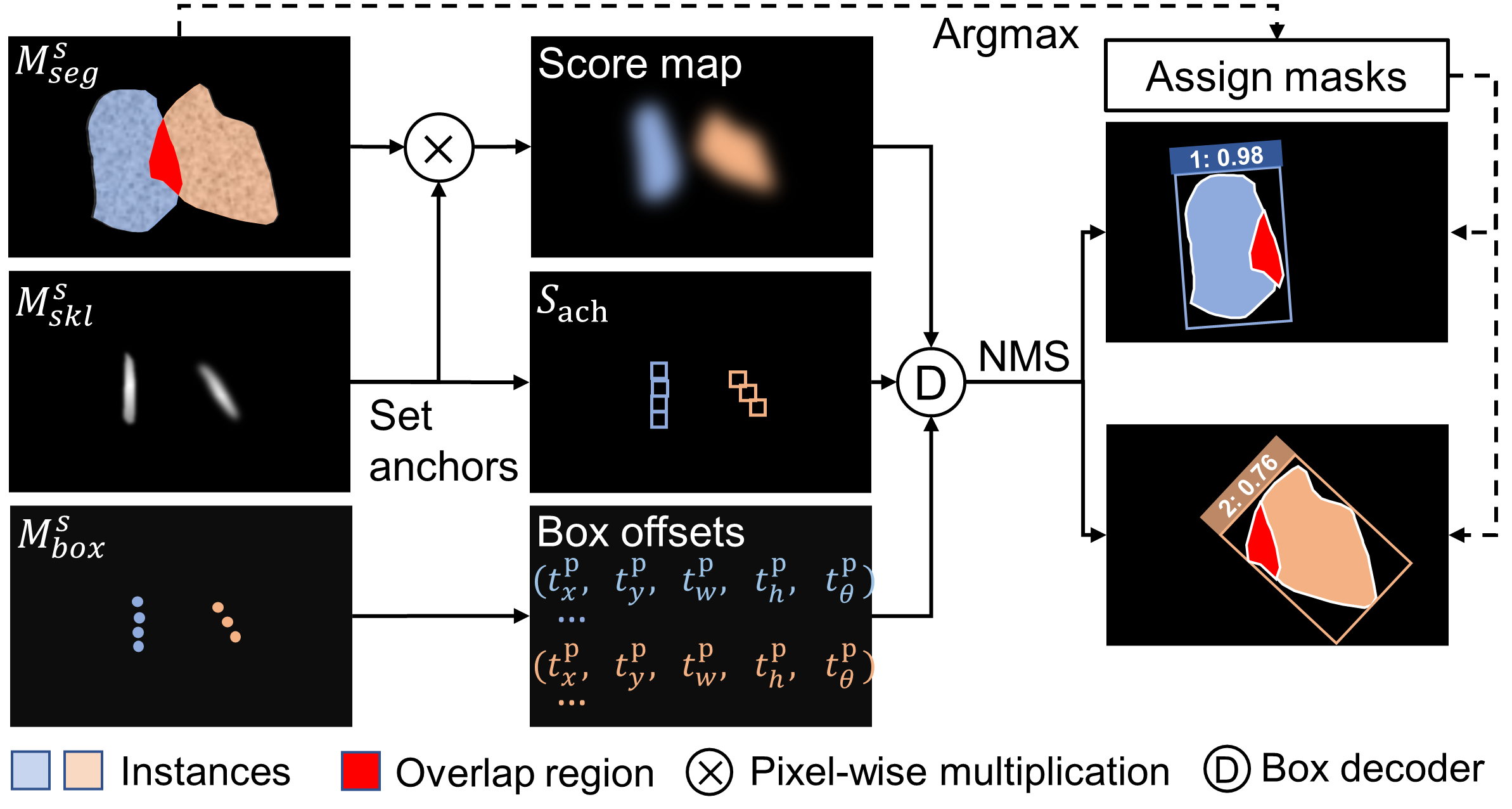}
    \caption{Illustration of box and mask proposal in the testing stage. The $\arg\max(M_{\text{seg}}^s)$  determines the masks (i.e., labeled regions) for predicted boxes. The boxes are obtained by decoding the anchors $S_{\text{ach}}$ based on the box offsets. To address the dense distribution issue, box scores are calculated from the pixel-wise multiplication of the segmentation map $M_{\text{seg}}^s$ and the skeleton map $M_{\text{skl}}^s$.}
\label{fig:6}
\end{figure*}

\subsection{Box and Mask Proposal}
\label{sec:box-mask-proposal}
Once the model is trained, we can obtain the rotated bounding boxes and their masks from the three head predictions as illustrated in Fig.~\ref{fig:6}. We first obtain instance masks (i.e., labeled regions) via calculating $\arg\max(M_{\text{seg}}^s)$ along the channels. Then, the anchors (denoted by the $S_{\text{ach}}$), set on the skeleton pixels, are decoded to bounding boxes based on the regressed box offsets. The box scores are obtained from the pixel-wise multiplication between the $M_{\text{seg}}^s$ and the skeleton map $M_{\text{skl}}^s$. This multiplication operation can reduce the confidence of boxes near the instance boundaries, since the skeleton map has higher values (maximum value of $1.0$) on the skeleton points of objects and lower values on other pixels. After all boxes are proposed, the rotated-NMS~\citep{citiation33} is performed to reduce the highly overlapped boxes. Finally, we reassign mask pixels to each box as follows: Only the overlapped region and the regions with the same classification type to the box are assigned. Noting that the box and mask proposal is only needed in the testing stage, which can avoid box-proposals during the training stage and significantly simplify the implementation.

\section{Experiments}
\subsection{Dataset}
Two large-scale datasets, namely \textit{PanNuke}~\citep{citiation14} and \textit{ChromSeg-SSL} are used to validate the proposed method. Table~\ref{tab:1} summarizes the splits for the model training and validation.

\textit{a) PanNuke}: This is a publicly available dataset consisting of over 7,000 histopathological patches collected from a local hospital and multiple public datasets, including Kumar~\citep{citiation34}, CMP2017~\citep{citiation35}, TCGA~\citep{citiation36}, and a dataset of bone marrow visual fields~\citep{citiation37}. The images correspond to $19$ different tissue types, and a total of $189,744$ nuclei were annotated and categorized into five clinically significant classes. In this study, we chose the PanNuke dataset as it contains enough samples for simulating SSL learning. We split the dataset into three subsets: the training set, the testing set, and the unlabeled set.

\textit{b) ChromSeg-SSL} We collected $4,185$ metaphase cell images with a resolution of $1600\times1600$ from the Obstetrics \& Gynecology Hospital of Fudan University. A total of $615$ images were annotated by five experienced cytologists using the LabelMe tool~\citep{citiation38}. The region of each chromosome was outlined with a polygon, and a label was given by the chromosome’s type (labels 1-22 for autosomes, 23 for X, and 24 for Y). In this study, all $3,570$ unlabeled images and $548$ labeled images were used for semi-supervised training. The remaining labeled images were used for testing. For further details regarding the data collection and annotation process, please refer to our data paper~\citep{citiation39}.

\begin{table}[tb]
\centering
\caption{The dataset for validating the proposed method}
\label{tab:1}
\begin{tabular}{lcccc}
\toprule
\textbf{Dataset} & \textbf{Samples} & \textbf{Train} & \textbf{Test} & \textbf{Unlabeled} \\ \midrule
\textit{PanNuke}  & Images & 1,901 & 718 & 4,656 \\
                  & Objects & 46,726 & 19,354 & - \\ \midrule
\textit{ChromSeg-SSL} & Images & 548 & 67 & 3,570 \\
                  & Objects & 23,356 & 2,869 & - \\ \bottomrule
\end{tabular}
\end{table}

\subsection{Implementation Details}
Four A2B-IS models with different sizes were trained in this study. Table~\ref{tab:2} summarizes the network architecture of the models, namely A2B-IS-B (Base size), A2B-IS-L (Large size), A2B-IS-S (Small size), and A2B-IS-T (Tiny size). All models have identical heads but different backbone networks. The input image size of the nuclei segmentation task and the chromosome segmentation task are $256\times256$  and $512\times512$, respectively. All models were trained using Google TensorFlow (version 2.8 with Keras API) on an NVIDIA RTX 3090Ti GPU with 24G memory. During the training stage, the multitask loss $\mathcal{L}_{\text{unlabeled}}+\lambda\mathcal{L}_{\text{labeled}}$ ($\lambda=4$ in this study) was minimized using the Adam optimizer with a learning rate of $0.0001$, decaying every epoch using an exponential rate of $0.96$. The number of epochs was $100$, and the batch size was $3$ (two labeled and one unlabeled image) and $2$ (one labeled and one unlabeled image) for the nuclei segmentation and the chromosome segmentation tasks, respectively. During the training, we conducted random flipping and rotation of images as data augmentation to enlarge the training set. 

\begin{table}[tb]
\centering
\caption{Four proposed A2B-IS models with different sizes. The $k$ and $s$ indicate the kernel size and the stride of Conv2d. The $C$ denotes channels of features. The $\times$ means the number of atrous self-attention modules in each A2B block. The asterisk * means a task-dependent value, i.e., $7$ and $26$ for the nuclei and the chromosome segmentation tasks, respectively.}
\label{tab:2}
\begin{tabular}{lcccc}
\toprule
\textbf{Layers} & \textbf{A2B-IS-B} & \textbf{A2B-IS-L} & \textbf{A2B-IS-S} & \textbf{A2B-IS-T} \\ \midrule
Stem & $C64, k3, s2$ & $C96, k3, s2$ & $C32, k3, s2$ & $C16, k3, s2$ \\ 
A2B \#1 & $C64, \times 4$ & $C96, \times 8$ & $C32, \times 4$ & $C16, \times 2$ \\ 
A2B \#2 & $C128, \times 4$ & $C192, \times 8$ & $C64, \times 4$ & $C32, \times 2$ \\
A2B \#3 & $C256, \times 4$ & $C384, \times 4$ & $C128, \times 4$ & $C64, \times 2$ \\ 
A2B \#4 & $C512, \times 4$ & $C768, \times 4$ & $C256, \times 4$ & $C128, \times 2$ \\ \midrule
\multirow{3}{*}{Seg head} & $C256, k3$ & $C256, k3$ & $C256, k3$ & $C256, k3$ \\ 
& $C256, k3$ & $C256, k3$ & $C256, k3$ & $C256, k3$ \\
& $C*, k3$ & $C*, k3$ & $C*, k3$ & $C*, k3$ \\ \midrule
\multirow{3}{*}{Skl head} & $C256, k3$ & $C256, k3$ & $C256, k3$ & $C256, k3$ \\ 
& $C256, k3$ & $C256, k3$ & $C256, k3$ & $C256, k3$ \\
& $C1, k3$ & $C1, k3$ & $C1, k3$ & $C1, k3$ \\ \midrule
\multirow{3}{*}{Box head} & $C256, k3$ & $C256, k3$ & $C256, k3$ & $C256, k3$ \\ 
& $C256, k3$ & $C256, k3$ & $C256, k3$ & $C256, k3$ \\
& $C5, k3$ & $C5, k3$ & $C5, k3$ & $C5, k3$ \\ 
\bottomrule
\end{tabular}
\end{table}

\subsection{Evaluation Metrics}
We adopted the following two standard metrics: the mean Average Precision ($mAP$)~\citep{citiation7} for detection and the mean Panoptic Quality ($mPQ$)~\citep{citiation14} for segmentation. The $mAP$ measures the mean Average Precision ($AP$) over all categories, and it was calculated when the $IoU$ threshold was set to $0.5$ for determining the true positives (i.e., the predicted boxes with $IoU \geq 0.5$ to any GT boxes). The $mPQ$ takes into account both the detection quality and the segmentation quality of all categories, making it especially suitable for assessing the performance of instance segmentation~\citep{citiation20}. Besides, to validate the performance on class-agnostic instance segmentation, the binary Panoptic Quality ($bPQ$) scores are also calculated for all methods.

\section{Results and discussion}

\begin{figure*}[!t]
    \centering
    \includegraphics[width=1.0\textwidth]{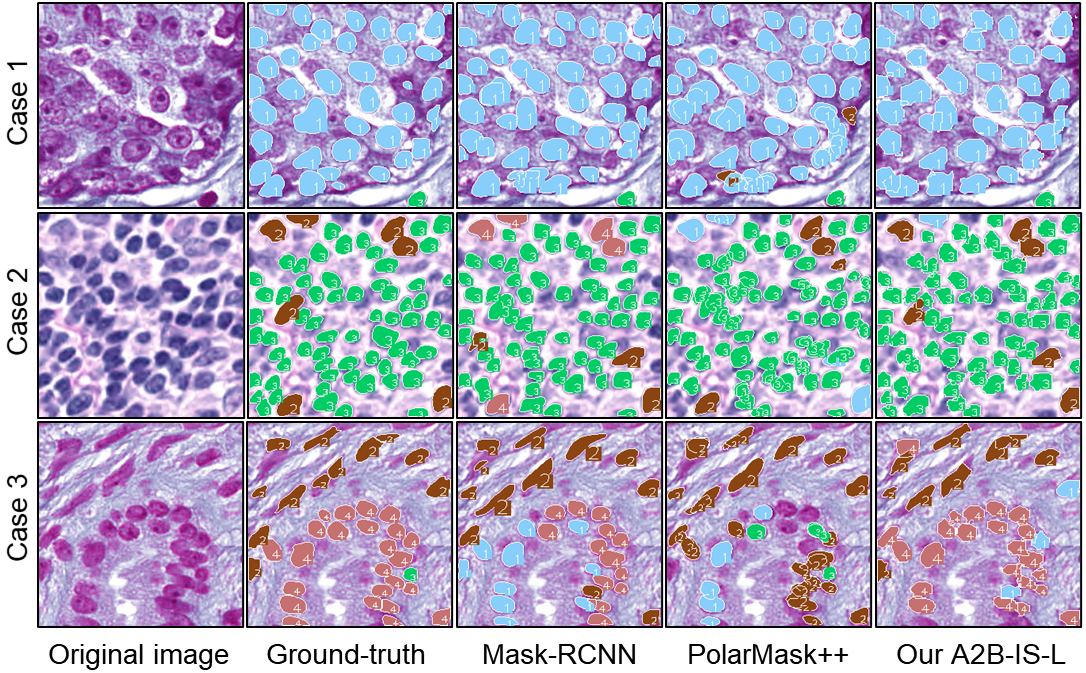}
    \caption{Result visualization of three nuclei instance segmentation cases. The white numbers indicate the class labels. For clarity, the predicted bounding boxes are not shown. The baseline Mask-RCNN and the SOTA single-stage method PolarMask++ have more misclassifications than our method (i.e., the A2B-IS-L), especially for case 2 and case 3.}
\label{fig:7}
\end{figure*}

\subsection{Comparison with SOTA IS methods}
To demonstrate the superiority of our A2B-IS for MS images, we compared it to the multi-stage method Mask-RCNN~\citep{citiation7} (baseline) and its variants, including the Cascade-RCNN~\citep{citiation15} (Casd-RCNN), HTC~\citep{citiation16}, SCNet~\citep{citiation17}, Mask-Scoring-RCNN (MS-RCNN)~\citep{citiation18}, and the QueryInst~\citep{citiation40}. We also evaluated three SOTA one-stage methods: the SOLOv2~\citep{citiation6}, the PolarMask++~\citep{citiation3}, and the SparseInst~\citep{citiation41}. These detection baselines were implemented using the ResNet-50-FPN backbone network in the MMDetection framework~\citep{citiation43}. For nuclei segmentation, we additionally retrained CellViT~\citep{citiation42} using the densely annotated PanNuke training images listed in Table~\ref{tab:1}. Besides, for the nuclei segmentation task, we also compared our method to the Hover-Net~\citep{citiation20} and TSFD-Net~\citep{citiation21} that tailored to the nuclei in histopathological images. These methods separate nuclei by predicting centroids along with contours or distance maps. To ensure a fair comparison, the semi-supervised learning strategy was not applied. All models were trained using the densely annotated training images listed in Table~\ref{tab:1}.

\begin{table}[htb]
\centering
\caption{Comparison with the SOTA instance segmentation methods on the \textit{PanNuke} dataset. The best performance is shown in bold. The SSL-IS is not applied to all models. }
\label{tab:3}
\begin{tabular}{lcccc}
\toprule
\multirow{2}{*}{\textbf{Methods}} & \textbf{Params} & \textbf{$mAP_{50}$} & \textbf{$mPQ$} & \textbf{$bPQ$} \\ 
& (MB) & (\%) & (\%) & (\%) \\ \midrule
Mask-RCNN~\citep{citiation7} & 166.9 & 48.3 (base) & 45.2 (base) & 63.9 (base) \\ 
Casd-RCNN~\citep{citiation15} & 293.0 & 48.7 (+0.4) & 43.7 (-1.5) & 63.5 (-0.4) \\ 
MS-RCNN~\citep{citiation18} & 228.9 & 48.0 (-0.3) & 44.8 (-0.4) & 63.8 (-0.1) \\ 
HTC~\citep{citiation16} & 293.6 & 47.9 (-0.4) & 44.3 (-0.9) & 63.7 (-0.2) \\ 
SCNet~\citep{citiation17} & 349.0 & 48.4 (+0.1) & 44.5 (-0.7) & 62.7 (-1.2) \\
QueryInst~\citep{citiation40} & 172.2 & 48.6 (+0.3) & 44.8 (-0.4) & 64.5 (+0.6) \\ \midrule
SOLOv2~\citep{citiation6} & 175.6 & 50.1 (+1.8) & 44.4 (-0.8) & 61.8 (-2.1) \\
PolarMask++~\citep{citiation3} & 130.8 & 46.8 (-1.5) & 45.7 (+0.5) & 59.1 (-4.8) \\
SparseInst~\citep{citiation41} & 32.7 & 43.3 (-5.0) & 40.1 (-5.1) & 56.3 (-7.6) \\ \midrule
HoVer-Net~\citep{citiation20} & 128.3 & 47.1 (-1.2) & 44.3 (-0.9) & 62.9 (-1.0) \\
TSFD-Net~\citep{citiation21} & 83.8 & 48.4 (+0.1) & 46.2 (+1.0) & 64.3 (+0.4) \\
CellViT~\citep{citiation42} & 186.7 & 49.3 (+1.0) & 46.5 (+1.3) & 65.1 (+1.2) \\ \midrule
A2B-IS-B & 87.7 & 49.2 (+0.9) & 45.5 (+0.3) & 64.1 (+0.2) \\
A2B-IS-L & 186.0 & 50.4 (+2.1) & 46.4 (+1.2) & 65.7 (+1.8) \\
A2B-IS-S & 30.6 & 47.6 (-0.7) & 44.6 (-0.6) & 61.4 (-2.5) \\ 
A2B-IS-T & 12.4 & 45.3 (-3.0) & 43.1 (-2.1) & 59.0 (-4.9) \\ \bottomrule
\end{tabular}
\end{table}

\textit{1) Nuclei segmentation performance:} Table \ref{tab:3} presents the results of the nuclei segmentation task, from which three main conclusions can be drawn. Firstly, segmenting nuclei in histopathological images from a diverse range of tissues is indeed a challenging task. The Mask-RCNN only achieves a $mAP_{50}$ score of $48.3\%$ and a $mPQ$ score of $45.2\%$. Interestingly, its variants with larger model sizes are even inferior (except the QueryInst). For example, the $mPQ$ score achieved by the SCNet is only $44.5\%$. Additionally, the $bPQ$ scores further support this conclusion. It can be observed that the gaps between the $bPQ$ scores and their corresponding $mPQ$ scores are large, reaching up to $20.0\%$. This phenomenon is primarily attributed to the fact that nuclei face severe issues of inter-class similarity. 

Secondly, it is evident that the one-stage methods, i.e., the SOLOv2, PolarMask++, and SparseInst, are significantly inferior to the multi-stage methods. For instance, the SparseInst only achieves a $mPQ$ score of $40.1\%$, which is much lower than Mask-RCNN’s $45.2\%$. Even the performance of the HoVer-Net specifically designed for nuclei segmentation is worse than the baseline version. In contrast, the cutting-edge method TSFD-Net demonstrates superior performance to the baseline, with higher scores for all $mAP$, $mPQ$, and $bPQ$ metrics. Our one-stage method A2B-IS-L further improves the performance, with a $mPQ$ score up to $46.4\%$. Even the smallest model, A2B-IS-T (only size of $12.4$ MB), can achieve good results with a $mPQ$ score of $43.1\%$.

Thirdly, our models have much smaller sizes than most existing multi-stage methods. For instance, the size of our A2B-IS-B model is only $87.7$MB, which is significantly smaller than the Mask-RCNN’s $166.9$MB. It is worth noting that a small model size is an essential advantage for teacher-student-based SSL, as it can substantially reduce the GPU memory requirements.

\begin{table}[htb]
\centering
\caption{Comparison with the SOTA instance segmentation methods on the \textit{ChromSeg-SSL} dataset. The best performance is shown in bold. The SSL-IS is not applied to all models. }
\label{tab:4}
\begin{tabular}{lcccc}
\toprule
\multirow{2}{*}{\textbf{Methods}} & \textbf{Params} & \textbf{$mAP_{50}$} & \textbf{$mPQ$} & \textbf{$bPQ$} \\ 
& (MB) & (\%) & (\%) & (\%) \\ \midrule
Mask-RCNN~\citep{citiation7} & 166.9 & 87.5 (base) & 83.8 (base) & 86.3 (base) \\
Casd-RCNN~\citep{citiation15} & 293.0 & 89.2 (+1.7) & 84.7 (+0.9) & 86.7 (+0.4)  \\ 
MS-RCNN~\citep{citiation18} & 228.9 & 87.4 (-0.1) & 84.2 (+0.4) & 86.1 (-0.2) \\ 
HTC~\citep{citiation16} & 293.6 & 89.8 (+2.3) & 85.0 (+1.2) & 86.8 (+0.5) \\ 
SCNet~\citep{citiation17} & 349.0 & 90.1 (+2.6) & 83.4 (-0.4) & 86.7 (+0.4) \\
QueryInst~\citep{citiation40} & 172.2 & 88.7 (+1.2) & 83.6 (-0.2) & 86.2 (-0.1) \\ \midrule
SOLOv2~\citep{citiation6} & 175.6 & 83.3 (-4.2) & 71.5 (-12.3) & 84.8 (-1.5) \\
PolarMask++~\citep{citiation3} & 130.8 & 82.7 (-4.8) & 70.0 (-13.8) & 79.6 (-6.7) \\
SparseInst~\citep{citiation41} & 32.7 & 81.6 (-5.9) & 69.3 (-14.5) & 78.4 (-7.9) \\ \midrule
A2B-IS-B & 87.7 & 91.9 (+4.4) & 84.4 (+0.6) & 86.5 (+0.2) \\
A2B-IS-L & 186.0 & 92.8 (+5.3) & 84.3 (+0.5) & 86.9 (+0.6) \\ 
A2B-IS-S  & 30.6 & 87.8 (+0.3) & 83.2 (-0.6) & 85.2 (-1.1) \\  
A2B-IS-T & 12.4 & 84.9 (-2.6) & 81.8 (-2.0) & 84.1 (-2.2) \\ 
\bottomrule
\end{tabular}
\end{table}

\begin{figure*}[!t]
    \centering
    \includegraphics[width=1.0\textwidth]{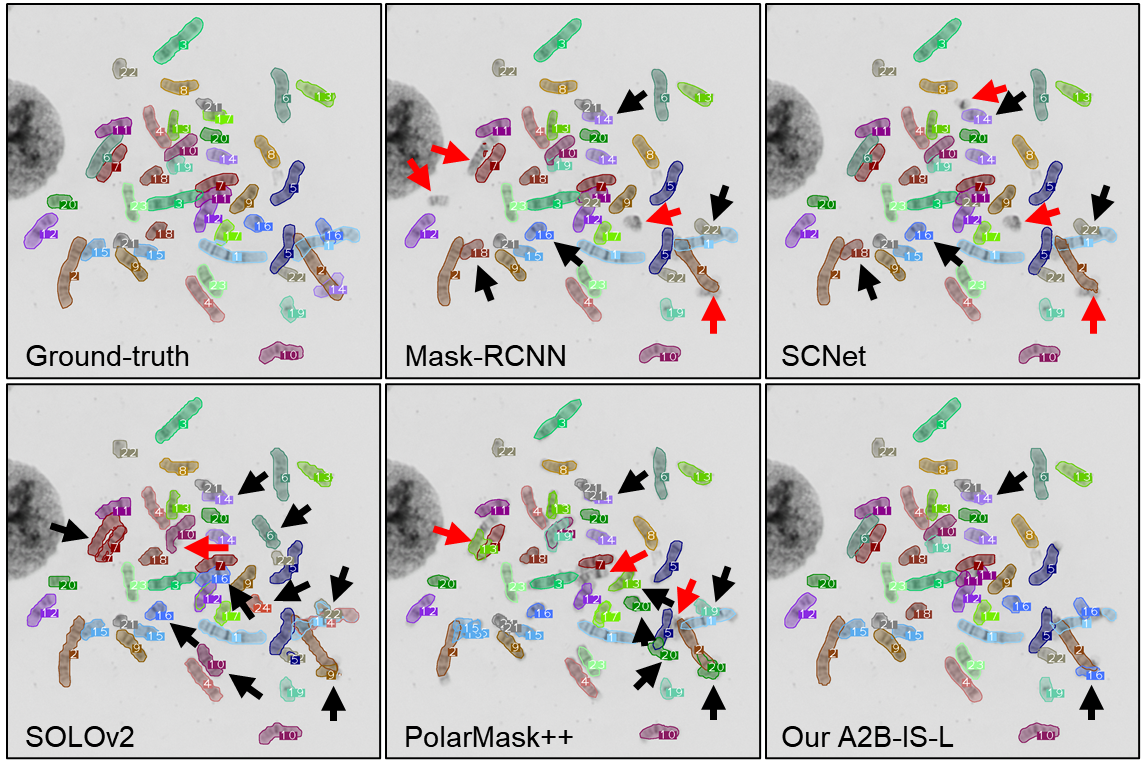}
    \caption{Result visualization of a chromosome instance segmentation case. The white numbers indicate the class labels. For clarity, the predicted bounding boxes are not shown. The red and the black arrows indicate false negatives and misclassifications, respectively. Existing methods have more misidentifications than our method (i.e., the A2B-IS-L).}
\label{fig:8}
\end{figure*}

\textit{2) Chromosome segmentation performance:} The results of chromosome segmentation are tabulated in Table \ref{tab:4}, which reveals similar conclusions to the nuclei segmentation task. However, compared to nuclei segmentation, the performance of our method is much more prominent in this task, whereas existing one-stage methods become much worse. For example, the $mAP$ and $mPQ$ scores achieved by our A2B-IS-L model are up to $92.8\%$ and $84.3\%$, respectively, with an improvement of $5.3\%$ and $0.5\%$ compared to the baseline. The counterparts achieved by the PolarMask++ are only $82.7\%$ and $70.0\%$. This is mainly because chromosomes in metaphase cell images exhibit much more complex features than nuclei, such as slender and bent shapes, dense distribution in arbitrary orientations, and even cross-overlapped. Existing methods face difficulties in handling these features.

\textit{3) Qualitative analysis of the performance:} We visualize some prediction results of the nuclei segmentation and the chromosome segmentation in Fig.~\ref{fig:7} and Fig.~\ref{fig:8}, respectively. It can be observed that existing methods have more misclassifications and false negatives than our method in both tasks. Besides, the masks predicted by existing single-stage methods (i.e., the SOLOV2 and PolarMask++) are visually worse than those of the multi-stage methods (i.e., the Mask-RCNN and the SCNet), especially in the chromosome segmentation task. In contrast, the mask quality of our method is close to that of the multi-stage methods.

\begin{figure*}[!t]
    \centering
    \includegraphics[width=1.0\textwidth]{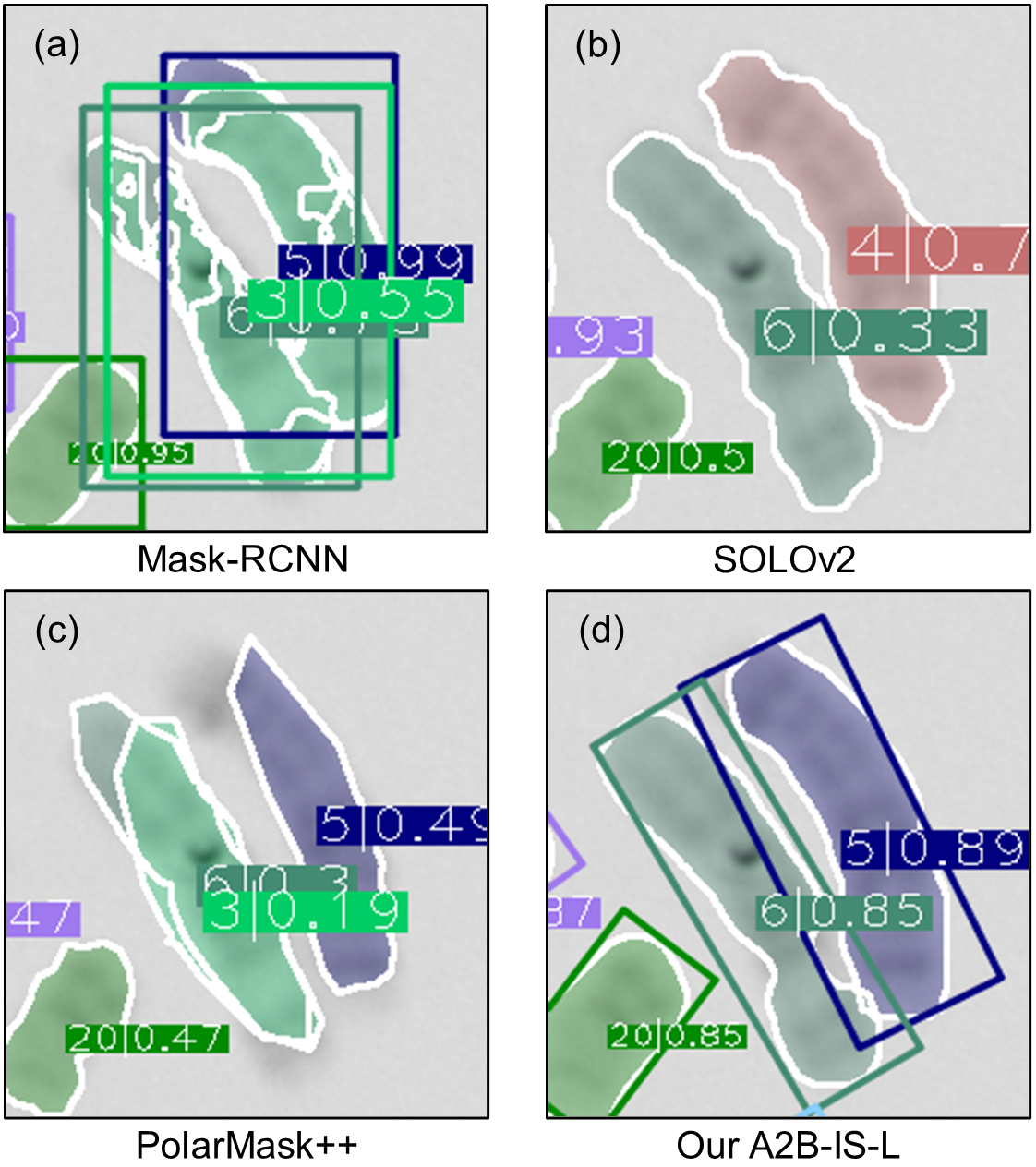}
    \caption{Local zoom images of results from Mask-RCNN, SOLOv2, PolarMask++, and our A2B-IS-L methods. Our method achieves the best segmentation masks.}
\label{fig:9}
\end{figure*}

We attribute the above phenomenon to the fact that existing box-based methods (i.e., the QueryInst, the Mask-RCNN and its variants) perform segmentation based on h-bbox proposals, which can hinder the model from discriminating the instances in cluster regions (see Fig.~\ref{fig:9}a for an example). Even though single-stage methods, e.g., the SOLOv2, the PolarMask++, and the SparseInst, have been developed to circumvent the box-proposals, they suffer from coarse segmentations as demonstrated by Fig.~\ref{fig:9}(b)-(c). This is mainly because they represent instances using grid cells, sampled contour points, or activation maps, which are not fine-grained enough for objects in MS images, especially for chromosomes with slender and bent shapes. In contrast, our A2B-IS method is the best one that represents objects with high-resolution pixel-wise masks and skeleton-guided r-bboxes (see Fig.~\ref{fig:9}d).

\subsection{Ablation Study}

\begin{table}[htb]
\centering
\small
\caption{The effectiveness of the ASA and the SSL-IS for training the A2B-IS models. The best improvement is shown in bold. ASA: Atrous Self-Attention; SSL-IS: Semi-Supervised Learning for Instance Segmentation. }
\label{tab:5}
\begin{tabular}{
    >{\raggedright\arraybackslash}p{1.9cm}|  
    >{\raggedright\arraybackslash}p{1.6cm}|  
    >{\centering\arraybackslash}p{1.1cm}|  
    >{\centering\arraybackslash}p{1.1cm}|  
    >{\centering\arraybackslash}p{1.1cm}|  
    >{\centering\arraybackslash}p{1.1cm}|  
    >{\centering\arraybackslash}p{1.1cm}|  
    >{\centering\arraybackslash}p{1.1cm}  
}
\toprule
\multirow{2}{*}{\textbf{Dataset}} & \multirow{2}{*}{\textbf{Strategy}} & \multicolumn{2}{c|}{\textbf{A2B-IS-B}} & \multicolumn{2}{c|}{\textbf{A2B-IS-S}} & \multicolumn{2}{c}{\textbf{A2B-IS-T}} \\
\cline{3-8}
& & $mAP_{50}$ & $mPQ$ & $mAP_{50}$ & $mPQ$ & $mAP_{50}$ & $mPQ$ \\
\midrule
\multirow{4}{1.9cm}{PanNuke} & \makecell[c]{\texttimes ASA \\ \texttimes SSL-IS} & \makecell[c]{44.3 \\ \scriptsize{(base)}} & \makecell[c]{42.5 \\ \scriptsize{(base)}} & \makecell[c]{42.2 \\ \scriptsize{(base)}} & \makecell[c]{41.0 \\ \scriptsize{(base)}} & \makecell[c]{37.9 \\ \scriptsize{(base)}} & \makecell[c]{39.7 \\ \scriptsize{(base)}} \\
\cline{2-8}
& \makecell[c]{\checkmark ASA \\ \texttimes SSL-IS} & \makecell[c]{49.2 \\ \scriptsize{(+4.9)}} & \makecell[c]{45.5 \\ \scriptsize{(+3.0)}} & \makecell[c]{47.6 \\ \scriptsize{(+5.4)}} & \makecell[c]{44.6 \\ \scriptsize{(+3.6)}} & \makecell[c]{45.3 \\ \scriptsize{(+7.4)}} & \makecell[c]{43.1 \\ \scriptsize{(+3.4)}} \\
\cline{2-8}
& \makecell[c]{\texttimes ASA \\ \checkmark SSL-IS} & \makecell[c]{45.3 \\ \scriptsize{(+1.0)}} & \makecell[c]{43.1 \\ \scriptsize{(+0.6)}} & \makecell[c]{43.0 \\ \scriptsize{(+0.8)}} & \makecell[c]{42.6 \\ \scriptsize{(+1.6)}} & \makecell[c]{38.0 \\ \scriptsize{(+0.1)}} & \makecell[c]{40.6 \\ \scriptsize{(+0.9)}} \\
\cline{2-8}
& \makecell[c]{\checkmark ASA \\ \checkmark SSL-IS} & \makecell[c]{\textbf{50.7} \\ \scriptsize{(+6.4)}} & \makecell[c]{\textbf{46.8} \\ \scriptsize{(+4.3)}} & \makecell[c]{\textbf{49.4} \\ \scriptsize{(+7.2)}} & \makecell[c]{\textbf{46.0} \\ \scriptsize{(+5.0)}} & \makecell[c]{\textbf{46.3} \\ \scriptsize{(+8.4)}} & \makecell[c]{\textbf{44.6} \\ \scriptsize{(+4.9)}} \\
\midrule
\multirow{4}{1.9cm}{ChromSeg-\allowbreak SSL} & \makecell[c]{\texttimes ASA \\ \texttimes SSL-IS} & \makecell[c]{89.7 \\ \scriptsize{(base)}} & \makecell[c]{83.1 \\ \scriptsize{(base)}} & \makecell[c]{85.4 \\ \scriptsize{(base)}} & \makecell[c]{80.7 \\ \scriptsize{(base)}} & \makecell[c]{75.0 \\ \scriptsize{(base)}} & \makecell[c]{75.3 \\ \scriptsize{(base)}} \\
\cline{2-8}
& \makecell[c]{\checkmark ASA \\ \texttimes SSL-IS} & \makecell[c]{91.9 \\ \scriptsize{(+2.2)}} & \makecell[c]{84.4 \\ \scriptsize{(+1.3)}} & \makecell[c]{87.8 \\ \scriptsize{(+2.4)}} & \makecell[c]{83.2 \\ \scriptsize{(+2.5)}} & \makecell[c]{84.9 \\ \scriptsize{(+9.9)}} & \makecell[c]{81.8 \\ \scriptsize{(+6.5)}} \\
\cline{2-8}
& \makecell[c]{\texttimes ASA \\ \checkmark SSL-IS} & \makecell[c]{90.7 \\ \scriptsize{(+1.0)}} & \makecell[c]{83.8 \\ \scriptsize{(+0.7)}} & \makecell[c]{86.7 \\ \scriptsize{(+1.3)}} & \makecell[c]{81.3 \\ \scriptsize{(+0.6)}} & \makecell[c]{77.9 \\ \scriptsize{(+2.9)}} & \makecell[c]{76.7 \\ \scriptsize{(+1.4)}} \\
\cline{2-8}
& \makecell[c]{\checkmark ASA \\ \checkmark SSL-IS} & \makecell[c]{\textbf{92.6} \\ \scriptsize{(+2.9)}} & \makecell[c]{\textbf{85.5} \\ \scriptsize{(+2.4)}} & \makecell[c]{\textbf{91.7} \\ \scriptsize{(+6.3)}} & \makecell[c]{\textbf{83.9} \\ \scriptsize{(+3.2)}} & \makecell[c]{\textbf{86.0} \\ \scriptsize{(+11.0)}} & \makecell[c]{\textbf{82.7} \\ \scriptsize{(+7.4)}} \\
\bottomrule
\end{tabular}
\end{table}

\begin{figure*}[!t]
    \centering
    \includegraphics[width=1.0\textwidth]{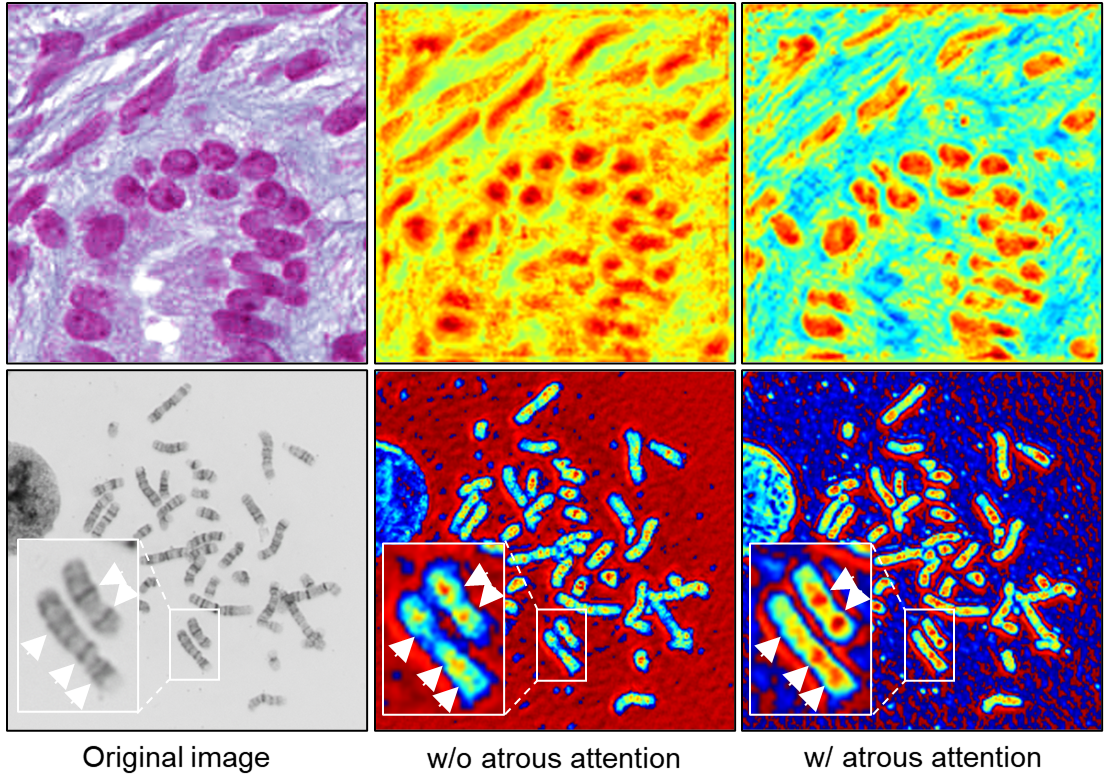}
    \caption{Feature visualization of a nuclei case and a chromosome case (the output feature maps of the last A2B block in the tiny model, i.e., the A2B\#4 in Table~\ref{tab:2}). Evidently, the model with the atrous attention pays more attention to the foreground regions to discriminate instances. The white arrows indicate the G-bands that are discriminative regions for distinguishing chromosome types.}
\label{fig:10}
\end{figure*}

\textit{1) The effectiveness of the Atrous Self-Attention (ASA) and the SSL-IS:} Two modules, i.e., the A2B backbone network with ASA and the SSL-IS are proposed to further improve the model performance. To quantify the effectiveness of these strategies, ablation studies are conducted by removing or adding the ASA and the SSL-IS strategies.

\begin{figure*}[!t]
    \centering
    \includegraphics[width=1.0\textwidth]{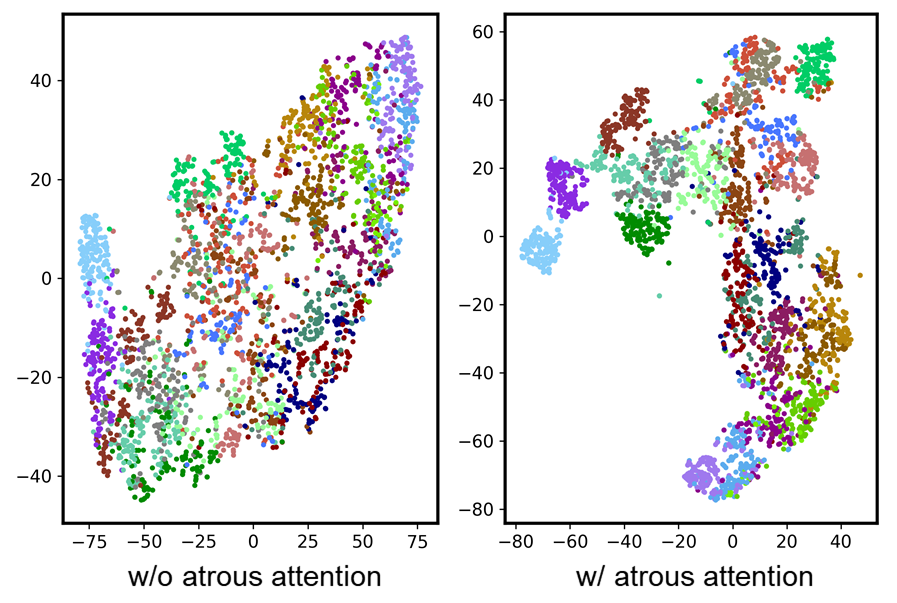}
    \caption{The t-SNE visualization of instance features extracted from the output feature maps of the last A2B block in the tiny model based on GT masks. The semantic features of the model with the atrous attention are more separatable than those without the attention.}
\label{fig:11}
\end{figure*}

Table~\ref{tab:5} shows the results. The ASA is vital for the model, as it significantly boosts the model's performance, especially for the tiny model. When only using the ASA, the improvements of the $mAP_{50}$ and the $mPQ$ of the A2B-IS-T are up to $7.4\%$ and $3.4\%$, respectively, for the nuclei segmentation task, while the counterparts in the chromosome segmentation task are $9.9\%$ and $6.5\%$. To analyze the effectiveness of the ASA qualitatively, we visualize the output feature maps of the last A2B block (i.e., the A2B\#4 in Table~\ref{tab:2}) in the tiny model. Fig.~\ref{fig:10} shows the visualization results of a nuclei case and a chromosome case. The model with the ASA pays more attention to the foreground regions. Furthermore, for statistical analysis, we extract the features of each instance according to their masks and conduct the t-SNE test. Fig.~\ref{fig:11} shows the t-SNE results of the chromosome segmentation task. Apparently, the semantic representations learned by the model with the ASA are more separatable than those of the model without the attention. 

The results in Table~\ref{tab:5} also demonstrate that the model performance can be further improved by applying the SSL-IS strategy for model training, especially when attention is also used. For instance, the $mPQ$ of the A2B-IS-T has increased from $39.7\%$ to $44.6\%$ in the nuclei segmentation task and from $75.3\%$ to $82.7\%$ in the chromosome segmentation task. The results demonstrate that the ASA mechanism can strengthen the model’s ability to mine valuable information from unlabeled data, thereby improving its overall performance.

\begin{figure*}[!t]
    \centering
    \includegraphics[width=1.0\textwidth]{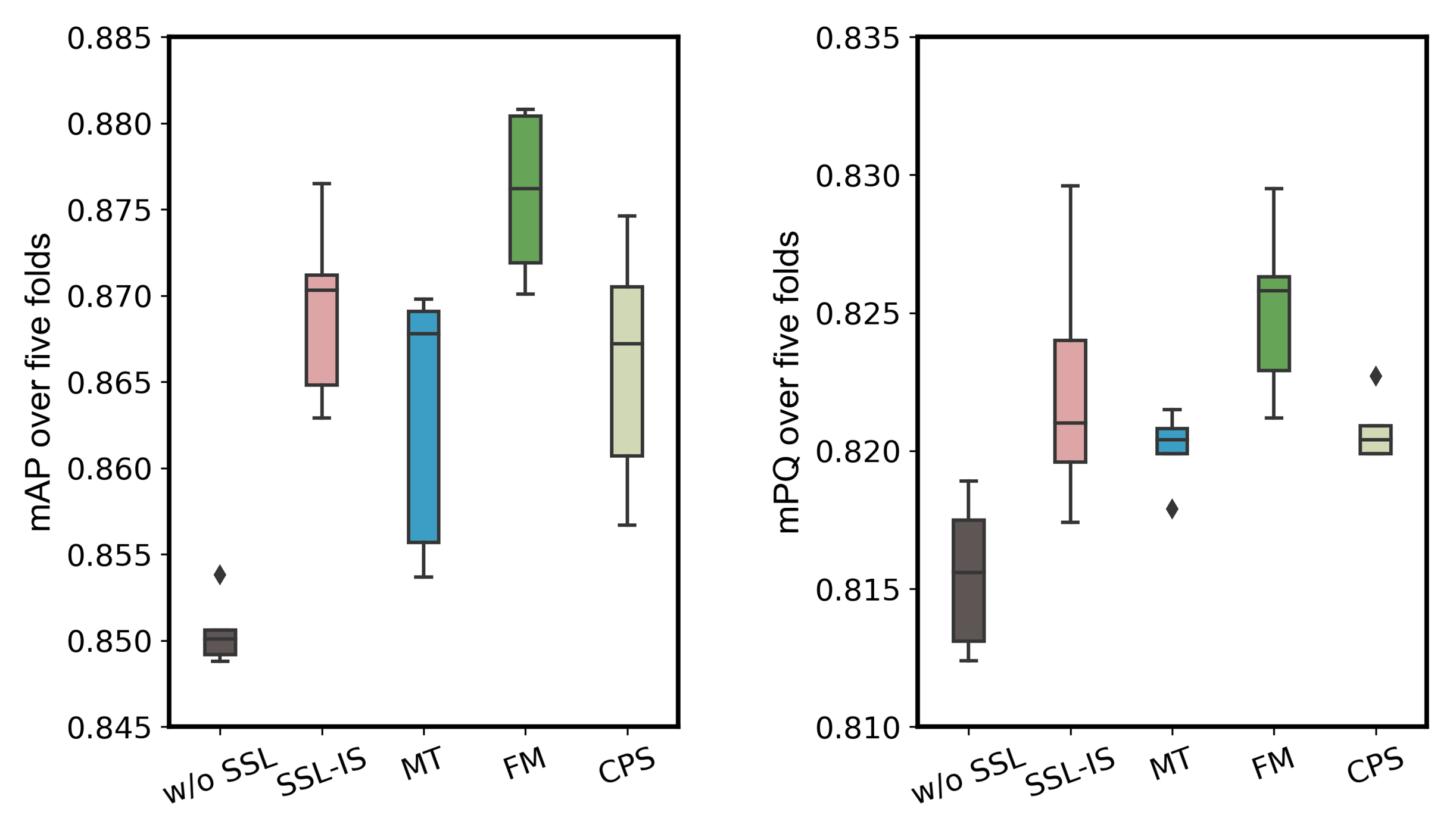}
    \caption{The 5-fold validation results demonstrate that all SSL methods can improve the model performance and achieve comparative performance.}
\label{fig:12}
\end{figure*}

\textit{2) Comparison with various SSL methods:} Based on our SSL-IS structure, we also adapt three SOTA SSL methods to our tasks, including the Mean-Teacher (MT)~\citep{citiation11}, the Fix-Match (FM)~\citep{citiation25}, and the Cross Pseudo Supervision (CPS)~\citep{citiation13}. Fig.~\ref{fig:12} shows the results of the chromosome segmentation task, which demonstrates that all the SSL methods can improve model performance by using unlabeled images for model training. All methods achieve comparative performance, but minor performance gaps remain. The FM method is the best one, followed by our SSL-IS method. However, a primary advantage of our method is that it requires fewer GPU memories than the other three methods. From Fig. 12, another interesting phenomenon is that the 5-fold $mAP$ scores of the SSL-IS and the FM method are more stable than those of the MT and the CPS methods, while it is contrary to the $mPQ$ scores. The cause of this phenomenon might be that the MT and the CPS methods also define pseudo-supervision loss in the labeled data. The pseudo-labels might contain wrong labels, especially in the pseudo-box-maps, which may slightly interfere with the detection performance.

\begin{figure*}[!t]
    \centering
    \includegraphics[width=1.0\textwidth]{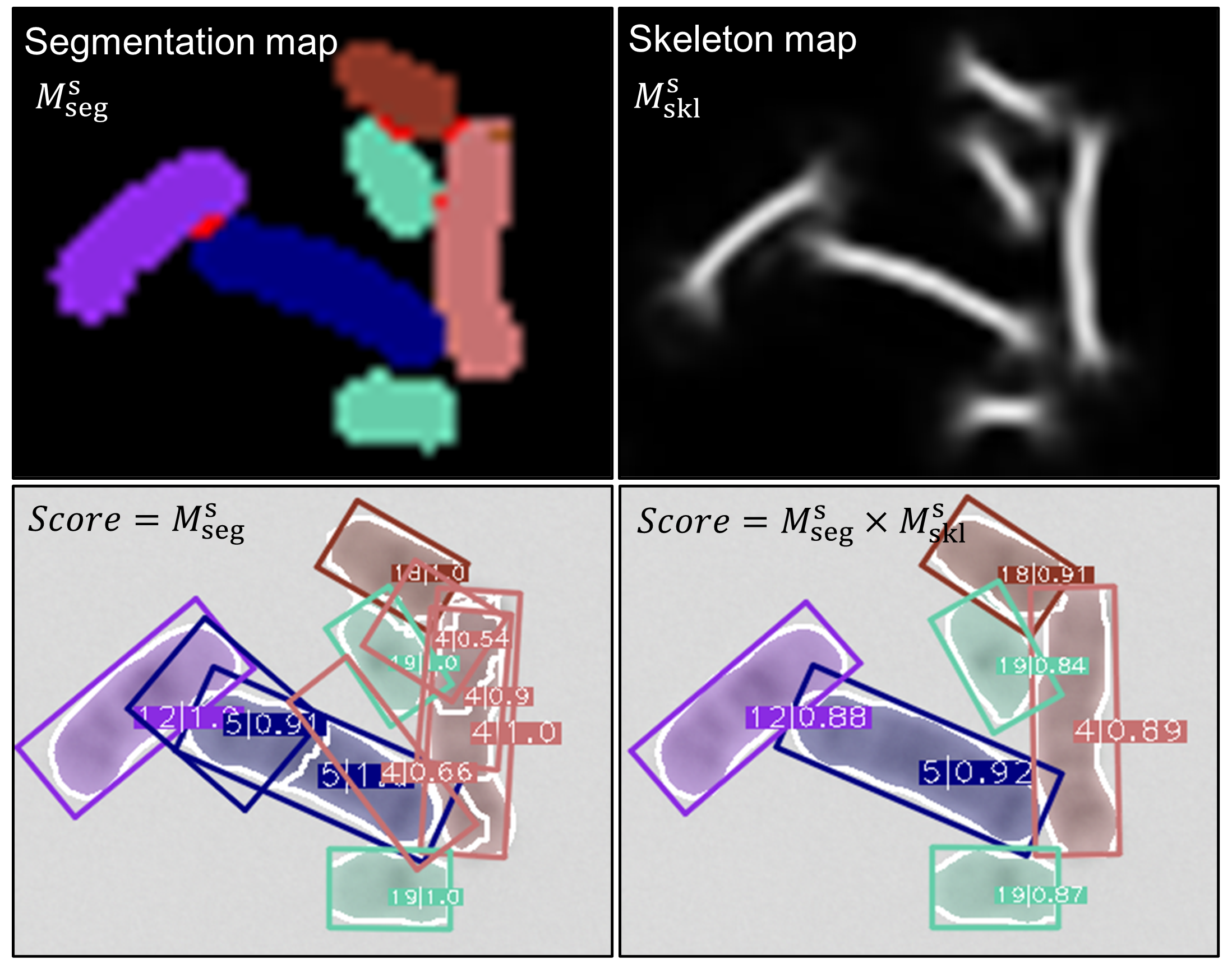}
    \caption{Prediction result of a chromosome segmentation case, which demonstrates that getting scores of box proposals from the multiplication of the segmentation map and the skeleton map can effectively address the dense distribution issue.}
\label{fig:13}
\end{figure*}

\textit{3) The skeleton map for addressing the dense distribution issue: } In the testing stage, to suppress the dense distribution issue, we multiply the skeleton map (i.e., the $M_{\text{skl}}^s$) to the segmentation map (i.e., the $M_{\text{seg}}^s$) to obtain the final scores for each box prediction. Here, we verify the effectiveness of this operation using Fig.~\ref{fig:13}, which visualizes the prediction results of a chromosome segmentation case. It can be observed that getting scores of box proposals from the multiplication of the segmentation map and the skeleton map can indeed address the dense distribution issue, with fewer bad box proposals near the instance boundaries.

\section{Conclusion}
In this paper, we proposed a detector named A2B-IS for accurate instance segmentation in cell microscopy images. To tackle the challenges posed by dense distribution, arbitrary orientations, and diverse shapes of objects, we represent instances using pixel-level classification masks and skeleton-guided rotated bounding boxes. To further simplify the model’s pipeline and enhance its representation ability, we designed the Atrous Attention Block to extract high-resolution feature maps. This novel design significantly facilitates semi-supervised learning, enabling efficient utilization of unlabeled images. Extensive experiments were performed to demonstrate the superiority of our method to most SOTA detectors. These studies yielded some attractive findings that are beneficial for both current applications and future research.

\section*{Code and data availability}
Code and dataset are available at \url{https://github.com/wangjuncongyu/A2B-Net}.

\section*{Acknowledgments}
This work was supported in part by the National Key Research and Development Program of China under Grant 2023YFC2705600 and 2023YFC2705601, and in part by the National Natural Science Foundation of China under Grant 62476237 and 6210131.



\bibliographystyle{elsarticle-harv} 
\bibliography{example}





\end{document}